\documentclass[runningheads]{llncs}
\usepackage[T1]{fontenc}
\usepackage{graphicx}
\usepackage{marvosym}
\usepackage{ifsym}
\usepackage{enumitem}
\usepackage{amssymb}

\usepackage{hyperref}

\begin{document}
\title{Feature Visualization in 3D Convolutional Neural Networks}
\author{Chunpeng Li\inst{1,2,3} \and
Ya-tang Li\inst{2,3(\textrm{\Letter})}}
\authorrunning{Chunpeng Li and Ya-tang Li}
\institute{College of Biological Sciences, China Agricultural University, Beijing, 100193, China \and
Beijing Institute for Brain Research, Chinese Academy of Medical Sciences, Peking Union Medical College, Beijing, 102206, China \and
Chinese Institute for Brain Research, Beijing 102206, China\\
\email{\{lichunpeng,yatangli\}@cibr.ac.cn}}
\maketitle              % typeset the header of the contribution
\begin{abstract}
Understanding the computations of convolutional neural networks requires effective visualization of their kernels. While maximal activation methods have proven successful in highlighting the preferred features of 2D convolutional kernels, directly applying these techniques to 3D convolutions often leads to uninterpretable results due to the higher dimensionality and complexity of 3D features. To address this challenge, we propose a novel visualization approach for 3D convolutional kernels that disentangles their texture and motion preferences. Our method begins with a data-driven decomposition of the optimal input that maximally activates a given kernel. We then introduce a two-stage optimization strategy to extract distinct texture and motion components from this input. Applying our approach to visualize kernels at various depths of several pre-trained models, we find that the resulting visualizations--particularly those capturing motion--clearly reveal the preferred dynamic patterns encoded by 3D kernels. These results demonstrate the effectiveness of our method in providing interpretable insights into 3D convolutional operations. Code is available at \url{https://github.com/YatangLiLab/3DKernelVisualizer}.

\end{abstract}
\section{Introduction}

With the rapid development of deep learning, neural network architectures have advanced significantly. Nevertheless, convolution remains a building block in most deep neural networks, especially in the field of computer vision. It serves as a powerful mechanism for extracting features from images and video data. Despite the success of end-to-end training on large datasets and the use of well-designed model architectures, the features learned by convolutional kernels often remain unclear. This lack of interpretability not only limits our understanding of how these networks function but also constrains our ability to design more effective models.

\begin{figure}
	\centering
	\includegraphics[width=0.9\textwidth]{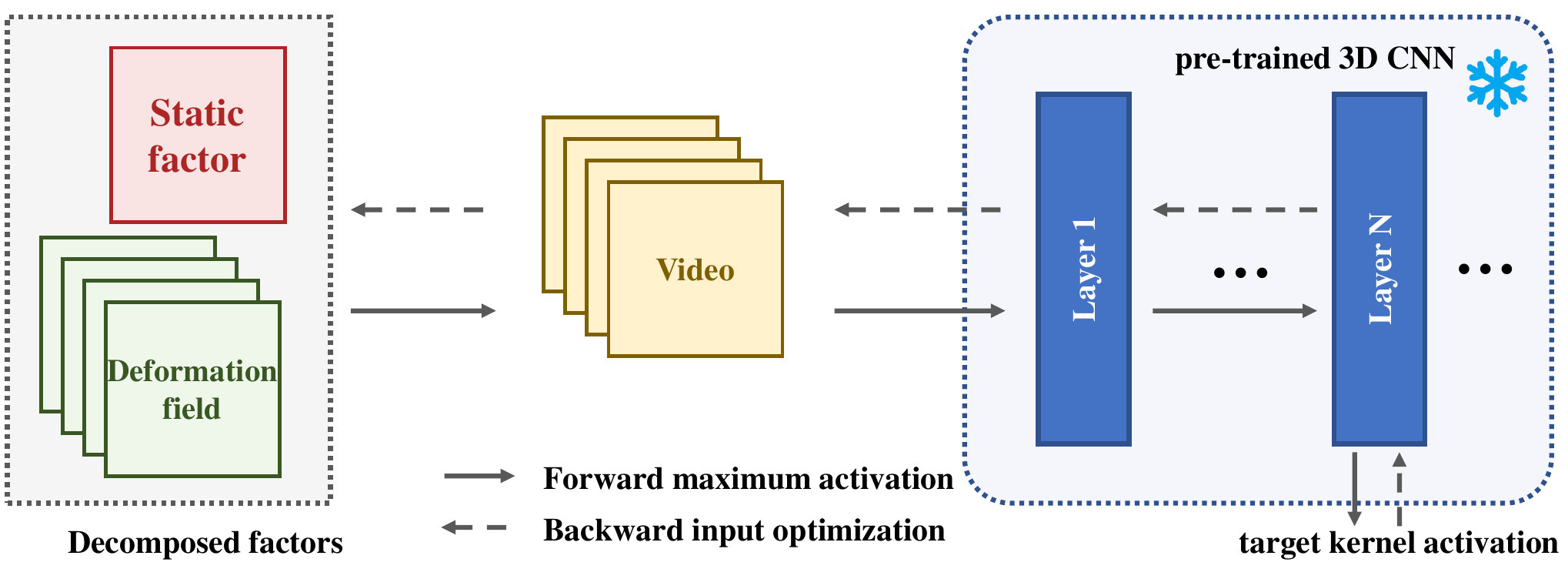}
	\caption{Overview of visualization process.} \label{fig1}
\end{figure}

Previous research has made significant progress in visualizing 2D convolutional kernels \cite{springenberg2014striving}, \cite{olah2017feature}, providing insights into how networks process static images. However, relatively little effort has been devoted to visualizing 3D convolutional kernels, despite their crucial role in video processing. Unlike 2D kernels, these 3D kernels capture spatial and temporal information simultaneously, which makes their visualization particularly challenging due to the need to interpret both spatial patterns and temporal dynamics.

A core difficulty in visualizing 3D kernels lies in effectively disentangling the static texture components from the motion-related features and in understanding their respective contributions to the convolutional activation. This complexity arises both from the coupling of spatial and temporal features and the high-dimensional nature of video data, which introduces significant noise and variability. Developing a visualization method that can decouple and independently present the texture and motion preferences of 3D kernels would not only deepen our understanding of the underlying computational mechanisms of convolution, but also inspire the design of more interpretable and efficient neural network architectures.

To address this challenge, we propose a novel visualization method that disentangles and separately presents the texture and motion preferences of 3D convolutional kernels. Our approach involves generating an optimal input that maximizes the activation of a given kernel, then decomposing this input into distinct texture and motion components using a data-driven representation. This decomposition allows us to isolate how each kernel responds to different types of features. To achieve this, we design a two-stage optimization procedure that ensures a clear separation of static and motion components. We then evaluate our method on a variety of 3D kernels from different layers of pre-trained models. Our results demonstrate the effectiveness of this visualization approach, providing insights into the spatiotemporal features learned by 3D convolutions and contributing to the broader goal of enhancing the interpretability of deep neural networks.

The contributions of this work are summarized as follows:

\begin{itemize}[leftmargin=2.5em]
	\item We introduce a data-driven method to decompose the optimal activation input of a 3D kernel into distinct texture and motion components.
	\item We propose a two-stage optimization strategy to effectively disentangle and independently visualize the static and dynamic features.
	\item We demonstrate the effectiveness of the proposed method through comprehensive experiments on multiple layers of pre-trained models.
\end{itemize}

\section{Related Work}
Three main lines of research have focused on the visualization and interpretation of 2D convolutional neural networks (CNNs). The first line involves gradient-based visualization techniques, which aim to identify the input features most critical for model predictions, thereby offering valuable insights into network behavior \cite{springenberg2014striving}, \cite{zhou2016learning}, \cite{selvaraju2017grad}, \cite{wang2020score}, \cite{jiang2021layercam}. However, these approaches often encounter difficulties in handling complex patterns and noise. The second line of research focuses on visualizing the kernels. This approach displays the raw weights of each kernel to reveal what has been learned. While intuitive, it lacks the context regarding how these kernels interact with input data \cite{NIPS2012_c399862d}, \cite{voss2021visualizing}. The third line employs activation maximization techniques to generate input patterns that maximize activation for specific kernels \cite{olah2017feature}, \cite{simonyan2013deep}, \cite{yosinski2015understanding}, though the resulting visualizations are often ambiguous and difficult to interpret. 

While these techniques have significantly advanced our understanding of 2D convolutional kernels, they fall short when applied to 3D convolutions. Unlike their 2D counterparts, 3D kernels encode both spatial (texture) and temporal (motion) information, introducing additional complexity. Prior attempts to extend 2D visualization methods directly to 3D kernels have yielded results that are difficult to interpret \cite{ras2020explainable}. This underscores the need for specialized visualization methods tailored to disentangle and independently represent the static and dynamic components inherent in 3D kernels. Our work addresses this gap by proposing a novel approach that explicitly decomposes the optimal activation input into separate texture and motion components, thereby extending the insights offered by existing 2D visualization techniques.

\section{Method}
The primary objective of our approach is to increase the interpretability of the optimal input that maximally activates individual 3D convolutional kernels, thereby revealing their feature preferences. To achieve this, we propose a data-driven decomposition method that visualizes the spatial and temporal preferences of 3D convolutional kernels, as well as the training strategy. As shown in Fig. \ref{fig1}, this method effectively disentangles texture and motion information to provide more intuitive visualizations.

\subsection{Motion Decomposition}
Directly extending 2D convolution visualization methods to 3D kernels fails to reveal the distinct motion preferences unique to 3D convolutions. While optical flow techniques offer a potential way of decomposing videos into motion and static components, they often lack robustness--particularly in scenarios involving random or non-coherent motion. These methods typically rely on local smoothness assumptions, making them ineffective when such assumptions are violated.

\begin{figure}
	\centering
	\includegraphics[width=0.7\textwidth]{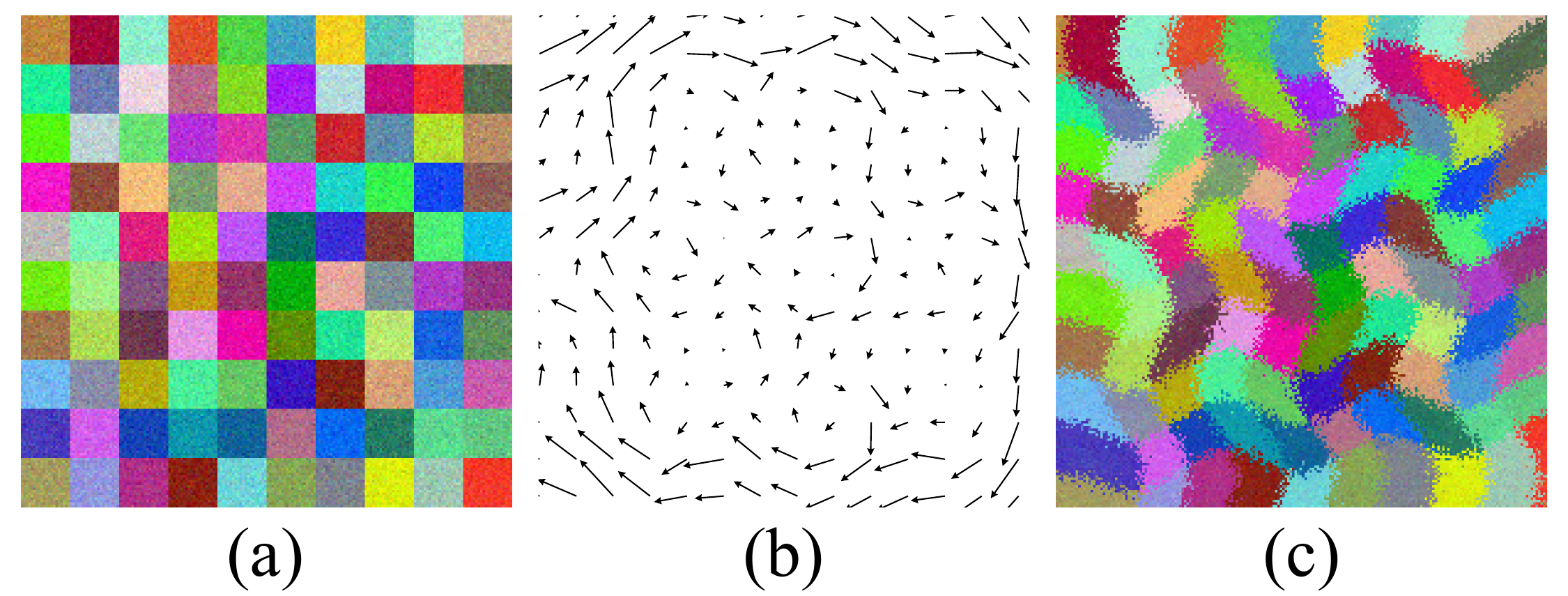}
	\caption{Illustration of the single frame reconstruction from decomposition factors: (a) Static factor $I$, (b) Deformation field $D_t$ at time $t$, (c) Frame $V_t$ at time t synthesized from (a) and (b).} \label{fig2}
\end{figure}

To overcome these limitations, we propose a data-driven approach for decomposing the visualizations of 3D convolutional kernels. Let $V=\{V_t\}_{t=1}^T$ denote the optimal input video sequence of $T$ frames. We decompose $V$ into two components:

\begin{equation}
	V_t=\mathcal{W}\left(I,D_t\right),\ \ \mathrm{for\ }t=1,2,\ldots,T,
\end{equation}
where:
\begin{itemize}[leftmargin=2.5em]
	\item $I \in \mathbb{R}^{H\times W\times C}$ is a learnable static factor representing a single frame that captures the kernel’s spatial (texture) preferences.
	\item $D_t \in \mathbb{R}^{H\times W\times2}$ is a learnable deformation field at time step $t$, which describes the pixel-wise displacement relative to the static image $I$. The two channels of $D_t$ correspond to horizontal and vertical displacements.
	
\end{itemize}
The warping operation $\mathcal{W}\left(I,D_t\right)$ generates the frame $V_t$ by spatially transforming the static image $I$ according to the deformation field $D_t$. For each pixel $p=\left(x,y\right)$ in $I$, its corresponding location in the frame at time t is computed as:

\begin{equation}
	p_t=\left(x+\mathrm{\Delta}x_t\left(x,y\right),  y+\mathrm{\Delta}y_t\left(x,y\right)\right),
\end{equation}
where $D_t\left(x,y\right)=\left(\mathrm{\Delta}x_t\left(x,y\right),\mathrm{\Delta}y_t\left(x,y\right)\right)$. Accordingly, the pixel value at location $p$ in $V_t$ is obtained by:
\begin{equation}
V_t\left(p\right)=I\left(p+D_t\left(p\right)\right).
\end{equation}
This equation performs a spatial transformation that is similar to deformable convolutions \cite{dcn} and optical flow estimation techniques \cite{raft}. It describes how each pixel in the video frames moves relative to the static component $I$, as illustrated in Fig. \ref{fig2}. This decomposition clearly separates the static and dynamic components in the video.

By operating over both spatial and temporal dimensions, this method ensures that the texture and motion preferences of a 3D convolutional kernel are independently represented. The static component $I$ encodes the texture preferences, while the dynamic component $D$ captures how these textures evolve over time. This decomposition facilitates a more interpretable and disentangled visualization of the kernel’s spatiotemporal features.

\subsection{Two-Stage Optimization}

Due to the complexity arising from the entanglement of spatial and temporal information, directly obtaining the learnable static and dynamic components through a single end-to-end optimization is challenging. To address this, we pro-pose a two-stage optimization strategy.

\subsubsection{Pixel-domain optimization.}
In the first stage, we optimize the input video in the pixel-domain to maximize the activation of a specific 3D convolutional kernel. This is driven by a conventional activation maximization loss. Let $f_k\left(V\right)$ denote the output feature map produced by passing video $V$ through kernel $k$, along with all preceding layers. The loss function for this stage is defined as:

\begin{equation}
	\mathcal{L}_{s1}=-\sum_{t=1}^{T}{f_k\left(V_t\right)}.
\end{equation}
This process generates a video input $V=\{V_t\}_{t=1}^T$ that maximally activate the target kernel, capturing both spatial and temporal features and laying a solid foundation for the subsequent decomposition.

\subsubsection{Decomposition-domain optimization.}
In the second stage, we decompose the optimized video $V$ into a static spatial component $I\in\mathbb{R}^{H\times W\times C}$ and a deformation field $D_t\in\mathbb{R}^{H\times W\times2}$. To ensure that these decomposed components faithfully represent the original video and yield meaningful interpretation, we design a composite loss function consisting of four terms:

\begin{enumerate}[label=(\arabic*), leftmargin=2.5em]
		\item \emph{Reconstruction Loss}.
		To ensure that $I$ and $D_t$ can faithfully reconstruct the video $V$ via a warping operation $\mathcal{W}\left(I,D_t\right)$, we use a mean squared error (MSE) loss:
		\begin{equation}
			\mathcal{L}_{recon}=\frac{1}{T}\sum_{t=1}^{T}\left \| V_t-\mathcal{W}\left(I,D_t\right) \right \| _2^2.
		\end{equation}
		
		\item \emph{Deformation Field Smoothness Loss}.
		To promote spatial smoothness in the deformation field, we apply a total variation (TV) regularization:
		\begin{equation}
			\mathcal{L}_{smooth_D}=\frac{1}{T}\sum_{t=1}^{T}\mathrm{TV}\left(D_t\right),
		\end{equation}
	
	\item \emph{Static Component Smoothness Loss}.
	Similarly, to prevent high-frequency noise in the static component I, we apply TV regularization:
	\begin{equation}
		\mathcal{L}_{smooth_I}=\mathrm{TV}\left(I\right).
	\end{equation}

	\item \emph{Static Component Similarity Loss}.
	To ensure that the static component $I$ captures the texture preference of the video, we promote similarity to the first frame $V_0$:
	\begin{equation}
		\mathcal{L}_{\mathrm{static}}=\left\|V_0-I\right\|_2^2.
	\end{equation}
	
\end{enumerate}
The total loss for the second stage is given by:
\begin{equation}
	\mathcal{L}_{s2}=\mathcal{L}_{recon}+\mathcal{L}_{smooth_D}+\mathcal{L}_{smooth_I}+\mathcal{L}_{static}.
\end{equation}
This combined loss ensures that the decomposed components not only reconstruct the video faithfully but also yield interpretable visualizations.

In summary, the proposed method enhances the interpretability of 3D convolutional kernels by decomposing their optimal input video into separate static and dynamic components through a two-stage optimization process. The first stage performs pixel-domain optimization to generate an input video that maximally activates the target kernel. The second stage decomposes the video into a static image $I$ and a set of deformation fields $D_t$ using a composite loss that ensures accurate reconstruction, smoothness, and similarity with the original input. This approach enables a clear and interpretable visualization of the spatiotemporal features encoded by 3D convolutional kernels, addressing the challenges posed by their high-dimensional structure. 

\section{Experiments}

In this section, we first describe the experimental setup. We then conduct ablation studies to evaluate the effectiveness of different loss components and the two-stage optimization strategy. Finally, we present comprehensive visualizations on widely used 3D CNN models, including C3D \cite{c3d} trained on Sports-1M \cite{sports1m}, I3D \cite{i3d} trained on Kinetics 400 \cite{kay2017kinetics} and 3D VQ-VAE \cite{opensora} trained on UCF-101 \cite{soomro2012ucf101}, to demonstrate the applicability and robustness of our proposed method.

\begin{figure}
	\centering
	\includegraphics[width=0.6\textwidth]{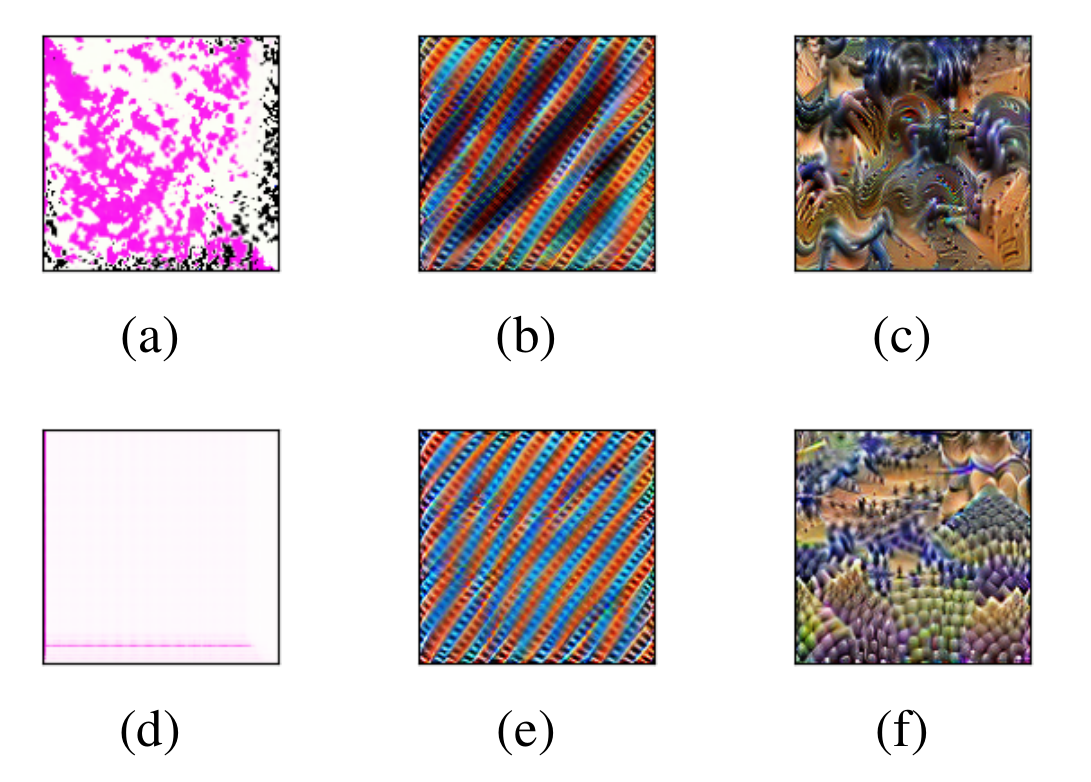}
	\caption{Visualization examples optimized in different domains. (a-c) are optimized in the Fourier domain, while (d-f) are in the pixel domain. Each column shows kernels at different layers. (a, d) are from I3D stage 1, (b, e) are from stage 3, and (c, f) are from stage 5.} \label{fig3}
\end{figure}

\subsection{Implementation Details}
In the first optimization stage, we use a pre-trained I3D \cite{i3d} with frozen weights. The initial video input is sampled from a normal distribution, with a spatial resolution of 128$\times$128, 16 frames, and 3 channels. A sigmoid function is applied to constrain input values within a valid range. We introduce random augmentations, including slight random shifts and scaling along the temporal, height, and width dimensions. Optimization is performed using the Adam optimizer with a learning rate of 0.05 for 1500 steps.

In the second stage, both the static image and deformation fields at each time step are initialized from a normal distribution. Video frames are reconstructed using PyTorch’s grid sampling function, with reconstruction error computed against the output from the first stage. These factors are optimized using Adam with a learning rate of 0.1 over 2000 iterations.

To visualize motion, we calculate the difference between adjacent deformation fields, revealing inter-frame shifts more clearly. For intuitive visualization, the deformation field is mapped to the HSV color space: hue encodes motion direction, while value represents motion magnitude. 

\subsection{Ablation Study}
We begin by evaluating the direct transfer of 2D visualization methods to the 3D Fourier domain. We then conduct a series of ablation studies to examine the contributions of individual loss components and the proposed two-stage optimization strategy. These experiments assess how each component affects the generation of interpretable visualizations for 3D convolutional kernels.

Due to the lack of ground-truth references for optimal input and deformation fields, traditional reference-based metrics like SSIM or PSNR are not applicable. While no-reference video quality evaluation methods \cite{nawala2017modeling}, \cite{de2024no} exist, they are designed for natural videos and thus fail to capture the characteristics of the optimized inputs, which often exhibit repetitive patterns and temporal jitter. Consequently, we rely solely on qualitative visual comparisons to evaluate the effectiveness of our method.

\begin{figure}[ht!]
	\centering
	\includegraphics[width=\textwidth]{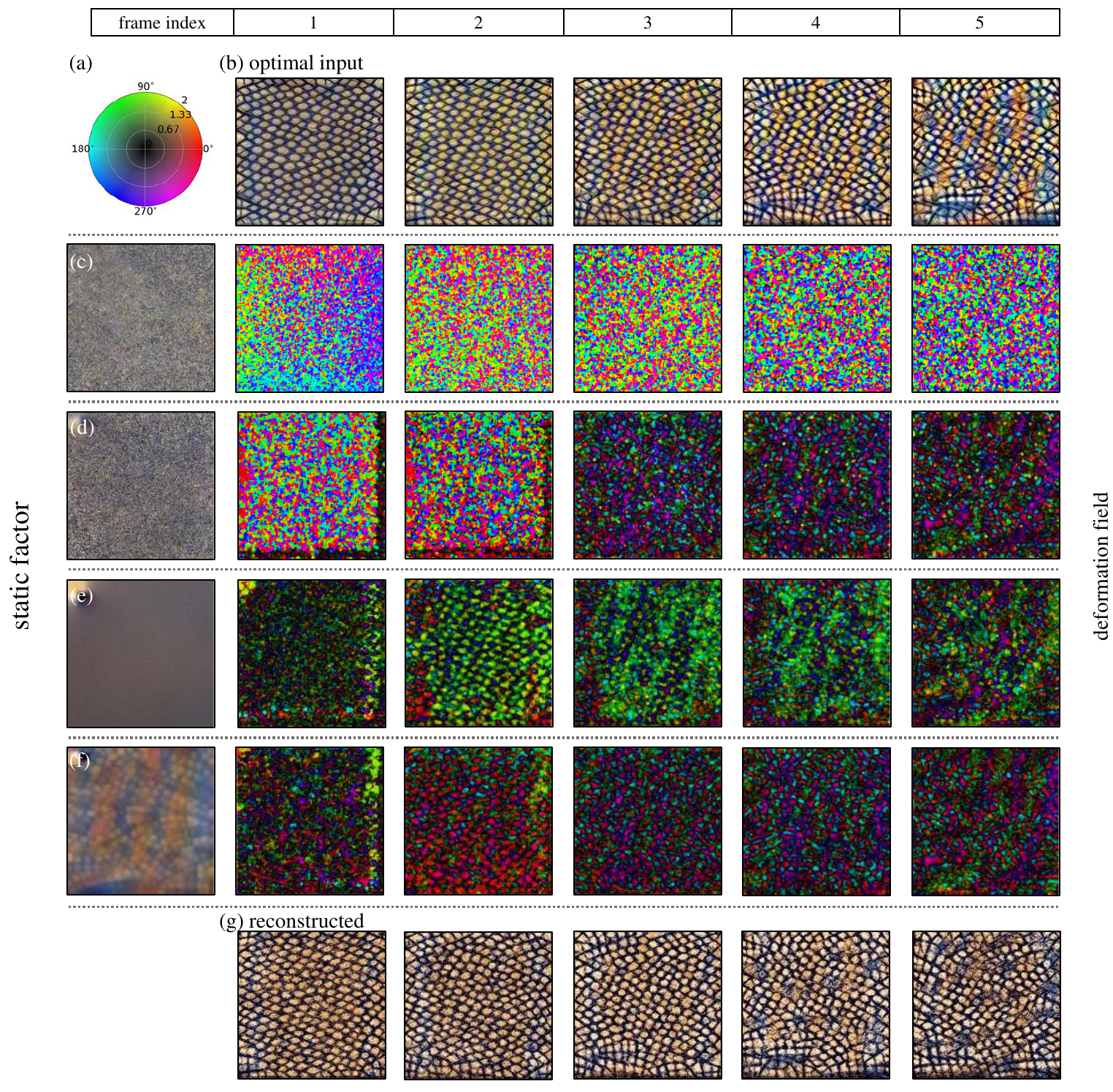}
	\caption{Ablation study on the loss function, using a kernel from I3D stage 4. (a) Color coding of the deformation field, where hue denotes the motion direction and value denotes the motion amplitude of each pixel. (b) The optimal input in the pixel domain. (c-f) Learnable components under different loss functions in the second training stage, where the first column is the static factor, and the last 3 columns are deformation fields at different times: (c) only includes reconstruction loss. (d) adds deformation TV loss, (e) adds static TV loss, (f) uses the full loss. (g) Reconstructed video with full loss.} \label{fig4}
\end{figure} 

\subsubsection{Ablation of Optimization Domain.}
We begin by directly extending the 2D convolution kernel visualization method in the Fourier domain to the 3D case. However, the amplitude spectrum in the 3D Fourier domain exhibits a significantly broader range. To address this in the 2D setting, a predefined scaling factor was introduced \cite{olah2017feature}, where the amplitude spectrum is computed as the product of a learnable parameter and a fixed scale, effectively constraining the parameter range. In the 3D setting, however, the amplitude spectrum spans a much wider range, and a predefined scale may not generalize well across different kernels. This can result in visual artifacts, as illustrated in Fig. \ref{fig3}. Therefore, direct optimization in the pixel domain is more suitable for our purpose.

\subsubsection{Ablation of Loss Function.}
We further explore the effect of individual loss components. As shown in Fig. \ref{fig4}, including consistency losses for both the static factor and deformation field significantly reduces noise. Without a meaningful static factor, however, it is difficult to interpret the deformation field. To address this, we introduce a similarity constraint to align the static factor with the first video frame, improving its semantic coherence. These results demonstrate the necessity of each loss component for generating accurate and interpretable visualizations.

\begin{figure}
	\centering
	\includegraphics[width=\textwidth]{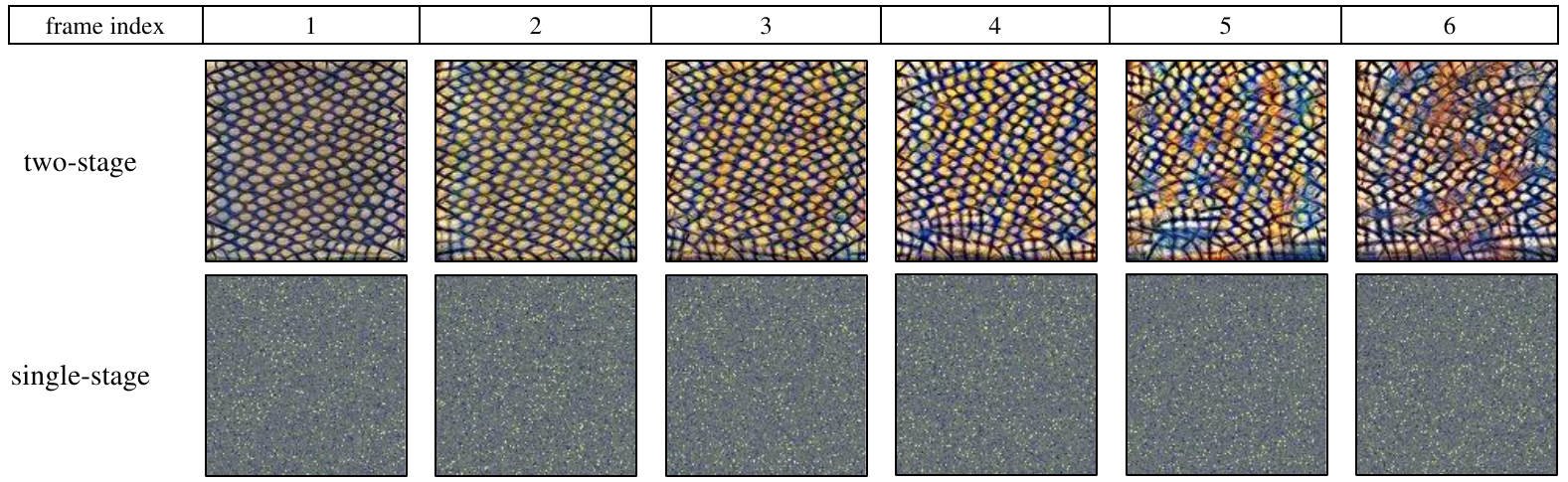}
	\caption{Ablation study on the optimization strategy. The first row shows pixel-domain visualizations from the two-stage method; the second row shows results from a single-stage approach.} \label{fig5}
\end{figure}

\subsubsection{Ablation of Training Strategy.}
We compare the proposed two-stage optimization with a single-stage approach in which static and dynamic components are learned simultaneously. As shown in Fig. \ref{fig5}, the single-stage approach often fails to converge to meaningful solutions. In contrast, the two-stage strategy improves convergence and leads to cleaner separation between texture and motion, confirming its effectiveness.

\subsection{Visualizations}
\begin{figure}[t]
	\centering
	\includegraphics[width=\textwidth]{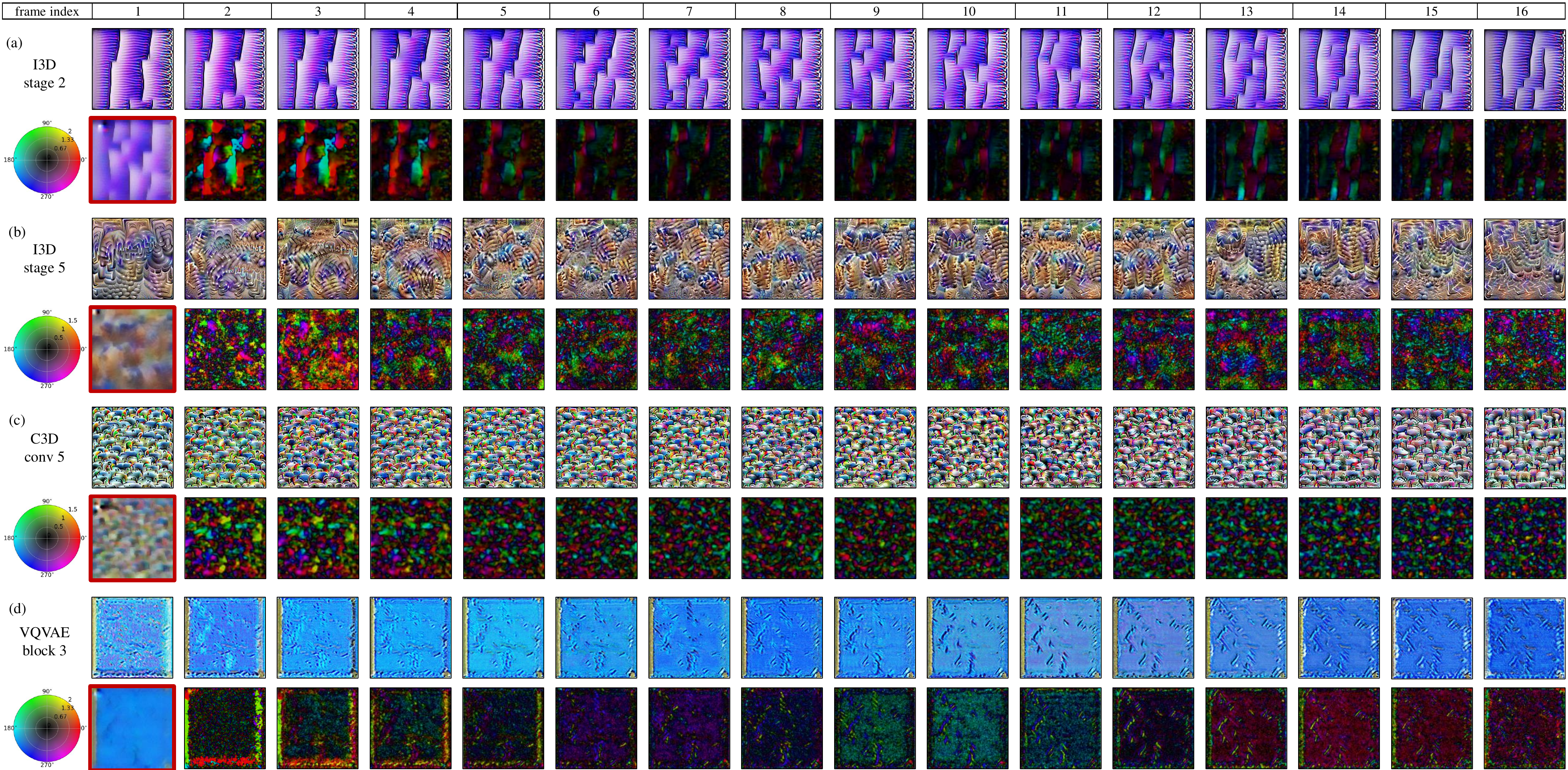}
	\caption{Visualization of representative kernels. (a-d) Kernels from different layers in 3 models. Each panel includes the optimal input video (top) and the decomposed static component and deformation fields (bottom row). } \label{fig6}
\end{figure}

In Fig. \ref{fig6}, different motion directions are color-coded to reveal the spatiotemporal preferences of each kernel. The results show that many 3D convolutional kernels prefer patch-wise textures that jitter in space, with interleaved patches moving in different directions.

Moreover, the observed motion is not simply a global shift. As model depth increases, motion patterns become increasingly complex. Different frames exhibit varying directions and amplitudes of motion, indicating that these kernels capture intricate, non-uniform spatiotemporal structures. This behavior resembles the functional diversity seen in the receptive fields of neurons in the mouse visual cortex \cite{yiyi2022selective}.

Overall, our method not only produces interpretable representations of texture and motion preferences but also uncovers the evolving complexity of 3D kernel responses, bridging the gap between artificial and biological visual systems.

\section{Conclusion}
We present a novel visualization method that effectively disentangles and visualizes the texture and motion preferences of 3D convolutional kernels. Our approach introduces a data-driven decomposition that separates the optimal input video of a convolutional kernel into a static component and a set of deformation fields. This method efficiently addresses the challenge of interpreting optimal inputs with complex and repetitive patterns. A two-stage optimization strategy further ensures that the decomposed components are learned efficiently and meaningfully.

Through extensive ablation studies and visualizations across CNN models, such as I3D \cite{i3d}, C3D \cite{c3d}, and 3D VQ-VAE \cite{opensora}, we validate the effectiveness and robustness of our method. The results reveal that the preferred motion pat-terns of 3D kernels often involve jittering textures, patterns reminiscent of neuronal activity in the biological visual system. Our method not only enhances interpretability but also offers valuable insights into how 3D convolutional kernels extract and represent spatiotemporal features.

\begin{credits}
\subsubsection{\ackname} Ya-tang Li is supported by the start-up fund from CIBR.

\subsubsection{\discintname}
The authors declare no competing interests.
\end{credits}
\bibliographystyle{splncs04}
\bibliography{main}

\end{document}